\documentclass[conference]{IEEEtran}
\usepackage{graphicx}
\usepackage{epstopdf}
\usepackage{verbatim}
\usepackage{amsmath}
\begin{document}
\title{A GPS Pseudorange Based Cooperative Vehicular Distance Measurement Technique}


\author{\IEEEauthorblockN{Daiqin Yang\IEEEauthorrefmark{1},
Fang Zhao\IEEEauthorrefmark{2},
Kai Liu\IEEEauthorrefmark{1},
Hock Beng Lim\IEEEauthorrefmark{1},
Emilio Frazzoli \IEEEauthorrefmark{3},
Daniela Rus \IEEEauthorrefmark{3}}
\IEEEauthorblockA{\IEEEauthorrefmark{1}Intelligent Systems Centre,
Nanyang Technological University, Singapore}
\IEEEauthorblockA{\IEEEauthorrefmark{2}Singapore-MIT Alliance for
Research and Technology, Singapore}
\IEEEauthorblockA{\IEEEauthorrefmark{3}Massachusetts Institute of Technology, USA}
}


\maketitle

\begin{abstract}
%\boldmath
Accurate vehicular localization is important for various cooperative vehicle safety (CVS) applications such as collision avoidance, turning assistant, etc. In this paper, we propose a cooperative vehicular distance measurement technique based on the sharing of GPS pseudorange measurements and a weighted least squares method. The classic double difference pseudorange solution, which was originally designed for high-end survey level GPS systems, is adapted to low-end navigation level GPS receivers for its wide availability in ground vehicles. The Carrier to Noise Ratio (CNR) of raw pseudorange measurements are taken into account for noise mitigation. We present a Dedicated Short Range Communications (DSRC) based mechanism to implement the exchange of pseudorange information among neighboring vehicles. As demonstrated in field tests, our proposed technique increases the accuracy of the distance measurement significantly compared with the distance obtained from the GPS fixes.
\end{abstract}

\section{Introduction}
Accurate positioning is important for various cooperative vehicle safety (CVS) applications such as collision avoidance, turning assistant, etc. To enable these applications, the lane level accuracy of vehicular positioning is required. However, current commercially available GPS devices, which are widely used in ground vehicles for navigation tasks, typically report tens of meters of positioning errors. Thus, it is difficult to recover the lane level relationship among neighbor vehicles, which is essential for most of the safety critical applications. For example, when a vehicle initiates an emergency brake to respond to an unexpected pedestrian, it will broadcast a notification message to all its neighbors. On receiving this message, it should be the vehicle which immediately follows the brake-initiator in the same lane to respond first, even though it may not necessarily be the nearest one. As vehicles move along the road, their trajectories conform to the multi-lane structure of the road. The unique yet important lane level relationship makes it imperative to achieve higher positioning accuracy for lane level topology recovery.

Current vehicular localization techniques can be categorized into two main classes - absolute position based and relative distance based. The techniques used to recover the absolute positions of moving vehicles are mostly based on trilateration of measured distances to known anchors, such as the cellular assisted localization \cite{Ref:song} and the Global Positioning System (GPS) \cite{Ref:kaplan}. Various augmentation techniques have been proposed to improve the positioning accuracy, including the Wide Area Augmentation System (WAAS), European Geostationary Navigation Overlay Service (EGNOS) and Differential GPS \cite{Ref:du}. However, these augmentation systems all require support from large-scale infrastructures, and some of them are not globally available. Besides these trilateration mechanisms, a fingerprint approach \cite{Ref:Bshara} is also popular in recovering positions of mobile objects using pattern recognition algorithms and statistical features of received wireless signals, such as from Wifi or cellular tower beacons. The accuracy of these techniques depends largely on the granularity of the training data set and the accuracy of system calibrations, which are time consuming and costly.

Some applications such as autonomous driving and collision avoidance depend more on the relative positions among vehicles, instead of their absolute positions. Representative techniques to measure the distance between moving vehicles include the Received Signal Strength (RSS) \cite{Ref:parker}\cite{Ref:alam2}, Time of Arrival (ToA), and Time Difference of Arrival (TDoA) \cite{Ref:obeid}. However, as stated in  \cite{Ref:alam}\cite{Ref:malaney}, these methods can hardly achieve the ranging accuracy required by most cooperative positioning applications. Alam et. al. proposed a distance measurement method for cooperative positioning without using radio range \cite{Ref:alam}. It reduces the average distance error to around five meters, but this is still not sufficient for lane level topology discovery.

In this paper, a weighted least squares pseudorange double difference algorithm is proposed to further increase the accuracy level of GPS based distance detection. The classic double difference pseudorange solution, which was originally designed for high-end survey level GPS systems, is adapted to low-end navigation level GPS receivers for its wide availability in ground vehicles. The Carrier to Noise Ratio (CNR) of raw pseudorange measurements are taken into account for noise mitigation. As demonstrated by field tests, our new algorithm greatly improves the accuracy of distance estimation and achieves average distance errors of around 3 meters. With the improved distance measurements among several vehicles, we can then use techniques such as trilateration to determine their relative positions.

The rest of this paper is organized as follows. In Section II, we present our weighted least squares pseudorange double difference algorithm for estimating distances between two vehicles. Performance evaluation is presented in Section III, and Section IV concludes the paper and outlines future work.

\section{Pseudorange Based Cooperative Distance Detection with Weighted Least Squares Method }
In this section, on the basis of the introduction of double difference pseudorange solution, we propose a weighted least square algorithm to mitigate the impact of the significant noises in pseudorange measurements of commercial GPS receivers. In addition, we discuss the DSRC system that enables the cooperative distance detection.

\subsection{Pseudorange Based Distance Detection}
In GPS system, GPS receivers recover their positions by trilateration of the measured distances to multiple visible satellites. The distance to each satellite is derived from an estimated time of transmission, and this distance is called the pseudorange. Several sources contribute to the errors in the pseudorange measurements. The aggregate impact of these error factors can introduce around 10 meters inaccuracy to the calculated GPS fixes. The pseudorange measurement between GPS receiver $a$ to satellite $i$ can be decomposed into \cite{Ref:kaplan}:

$$PR^i_a = R^i_a + t_a + x^i + \varepsilon^i_a$$

\noindent where $R^i_a$ is the true distance between satellite $i$ and receiver $a$; $t_a$ is the bias caused by receiver $a$'s clock bias; $x^i$ is the common noise related to satellite $i$ that are shared by all the GPS receivers within a  vicinity region, including the satellite clock bias, the atmospheric delay, and the error in the broadcasted ephemeris; and $\varepsilon^i_a$ is the non-common noise specific to receiver $a$ and satellite $i$, including the multipath error and the acquisition noise. By taking the difference between the pseudoranges of two receivers $a$ and $b$ to the same satellite $i$, the common noise due to satellite $i$ can be effectively removed:

\begin{eqnarray}
S^i_{ab} &=& PR^i_a - PR^i_b  \nonumber \\
        &=& (R^i_a - R^i_b) + (t_a - t_b)+ (\varepsilon^i_a - \varepsilon^i_b)   \nonumber   \\
        &=& \Delta R^i_{ab} + (t_a - t_b)+ (\varepsilon^i_a - \varepsilon^i_b)
\end{eqnarray}

\noindent where $S^i_{ab}$ is the single difference of pseudorange measurements, and $\Delta R^i_{ab}$ is the difference between the true ranges from receiver $a$ and $b$ to satellite $i$. As the true ranges from satellite $i$ to GPS receivers $a$ and $b$ are much larger than the distance between $a$ and $b$, the two vectors pointing from $i$ to $a$ and $b$ are nearly parallel to each other. As illustrated in Figure 1, $\Delta R^i_{ab}$ can thus be approximated by:

$$\Delta R^i_{ab} = \vec{e}^i \cdot \vec{r}_{ab}$$

\noindent where $\vec{e}^i$ is the unit vector pointing from GPS receiver $a$ (or $b$) to satellite $i$, and $\vec{r}_{ab}$ is the distance vector between receiver $a$ and $b$. When double difference is used, the clock bias of receiver $a$ and $b$ can be further removed.

\begin{eqnarray}
D^{ij}_{ab}    &=& (PR^i_a - PR^i_b)-( PR^j_a - PR^j_b)   \nonumber \\
                &=& [\Delta R^i_{ab} - \Delta R^j_{ab}]+ [(\varepsilon^i_a - \varepsilon^i_b)- (\varepsilon^j_a - \varepsilon^j_b)]   \nonumber \\
                &=& [\vec{e}^i - \vec{e}^j]\cdot \vec{r}_{ab} + [(\varepsilon^i_a - \varepsilon^i_b)- (\varepsilon^j_a - \varepsilon^j_b)]
\label{eq:dd}
\end{eqnarray}

Given known position of either one of the GPS receiver, the unit vector of each $\vec{e}^i$ can be calculated using the received ephemeris of each target satellite $i$. Let $\mathbf{D}_{ab}=[D^{10}_{ab} \; D^{20}_{ab} \; \cdots \; D^{n0}_{ab}]^T$ denote the column vector of pseudorange double differences, and there are totally $n+1$ satellites shared by receiver $a$ and $b$. Eq. \ref{eq:dd} can be reorganized into:

\begin{eqnarray}
\mathbf{D}_{ab}= \mathbf{H} \vec{r}_{ab} + \boldsymbol{\epsilon}
\label{eq:dd_vec}
\end{eqnarray}

\noindent where $\mathbf{H} = [(\vec{e}^1 - \vec{e}^0) \; (\vec{e}^2 - \vec{e}^0) \; \cdots \; (\vec{e}^n - \vec{e}^0)]^T$, and $\boldsymbol{\epsilon}$ is the column vector of aggregated non-common noises related to the two satellites and two receivers involved. If $\boldsymbol{\epsilon}$ can be assumed to be zero mean and equal variance, $\vec{r}_{ab}$ can be derived by the linear least squares estimator

\begin{eqnarray}
\vec{r}_{ab}= (\mathbf{H}^T \mathbf{H})^{-1}  \mathbf{H}^T  \mathbf{D}_{ab}
\end{eqnarray}

\noindent as long as there are at least four shared satellites between the two GPS receivers.

Comparing with the around twenty thousand kilometers distances from satellites to vehicular GPS receivers, tens of meters positioning error to the GPS receiver $a$ or $b$ can be fairly ignored when calculating the unit vectors of $\vec{e}^i$. Thus, $\vec{e}^i$ can be calculated by directly using the final output of the GPS fix of either one of the two GPS receiver modules.

\begin{figure}[htb]
\centering
\includegraphics[width=2.8in, height=1.6in]{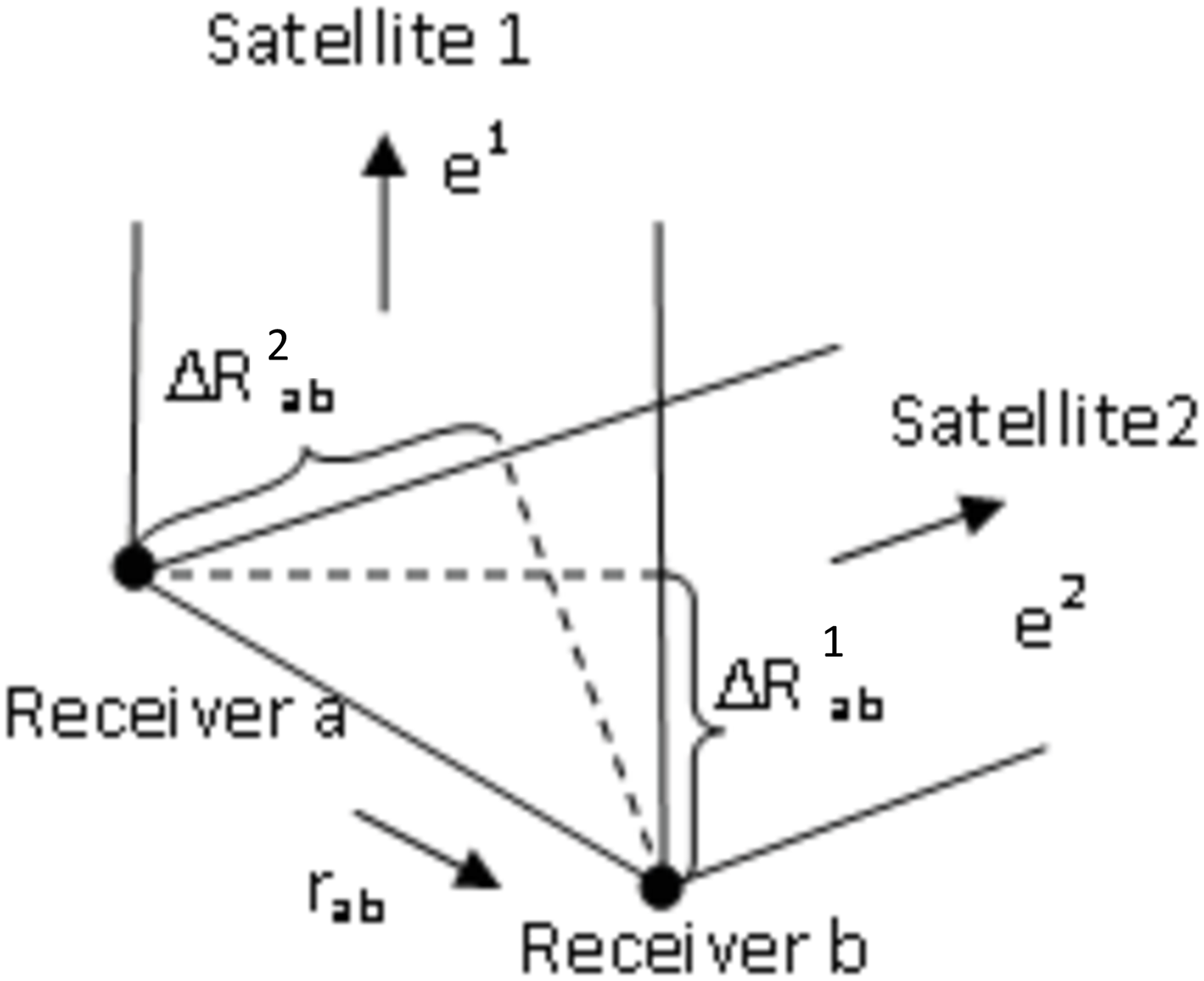}
\caption{Pseudorange double difference}
\label{fig:double_difference}
\end{figure}

%\begin{figure}[!t]
%\centering
%\includegraphics[width=2.5in]{dq.eps}
%\caption{Simulation Results}
%\label{fig_sim}
%\end{figure}

\subsection{Weighted Least Squares Approach}

In eq. \ref{eq:dd_vec}, $\boldsymbol{\epsilon}$ is the aggregated non-common noises of pseudorange measurements including both the multipath and random code acquisition errors encountered by the two GPS receivers. It's $i$th item is equal to

\begin{eqnarray}
\epsilon^i = (\varepsilon^i_a - \varepsilon^i_b) - (\varepsilon^0_a - \varepsilon^0_b)
\label{eq:noise}
\end{eqnarray}

In scenarios where multipath effect is not severe, for example when vehicles are moving along highways, $\mathbf{\epsilon}$ are dominated by code acquisition errors, which can be directly linked to the Carrier to Noise Ratio (CNR) of the received satellite signal. With commercial and navigation level GPS receivers, the accuracy of code acquisition is much worse than survey level devices. It can introduce more than ten meters of errors to the pseudorange measurements and thus severely degrade the accuracy of double difference based distance detections.

\begin{table*}[!t]
\renewcommand{\arraystretch}{1.3}
% if using array.sty, it might be a good idea to tweak the value of
% \extrarowheight as needed to properly center the text within the cells
\caption{The impact of Carrier to Noise Ratio to double difference based distance detection}
\label{table_example}
\centering
\begin{tabular}{|c|c||c|c|c|c|c|c|c|c|c|c|}
\hline
 & CNR Threshold (dBHz) & 47 & 46 & 45 & 44 & 43 & 42 & 41 & 40 & 35 & 30 \\
\hline
\hline
Baseline distance & Average Distance Error (m) & 0.983 & 3.049 & 4.241 & 4.286 & 4.301 & 4.868 & 4.492 & 4.384 & 4.429 & 5.265 \\
\cline{2-12}
of 3 meters & Number of Valid Samples & 14 & 362 & 2339 & 3736 & 4212 & 4713 & 5056 & 5175 & 5503 & 5516 \\
\hline
\hline
Baseline distance & Average Distance Error (m) & 0.030 & 4.3318 & 5.170 & 5.487 & 5.756 & 5.708 & 5.773 & 5.394 & 4.124 & 5.174 \\
\cline{2-12}
of 8 meters & Number of Valid Samples & 10 & 636 & 2529 & 4137 & 4794 & 5233 & 5414 & 5567 & 5711 & 5769 \\
\hline
\end{tabular}
\end{table*}

Table \ref{table_example} gives the experimental results of filtering pseudorange data with a CNR threshold. It shows the relationship among the accuracy of the calculated distances using normal least squares algorithm, the CNR threshold for pseudorange measurements, and the number of valid samples. Detailed settings of the experiment are elaborated in Section \ref{sec:experiment}. A set of CNR thresholds are used to choose valid pseudorange measurements for least squares based double difference processing. The satellite with the best CNRs to both of the  GPS receivers is selected as the reference. As shown in the table, a higher CNR threshold gives higher accuracy of distance detection. However, as it is not always possible to get enough (four or more) pseudorange measurements with high CNRs, the number of valid samples decreases accordingly, which implies substantial delays between successive distance measurements.

To address this problem, a CNR based weighted least squares method is adopted to takes into account the different accuracy level of different pseudorange measurements in estimating the distance. With non-equal variances of each $\epsilon^i$, the weighted least squares estimator

\begin{eqnarray}
\vec{r}_{ab}= (\mathbf{H}^T \mathbf{W} \mathbf{H})^{-1}  \mathbf{H}^T \mathbf{W} \mathbf{D}_{ab}
\label{eq:wls}
\end{eqnarray}

\noindent can be used as the best linear unbiased estimator (BLUE), where the weight matrix $\mathbf{W}$ is the inverse of the covariance matrix of $\mathbf{\epsilon}$. The following strategy is proposed to implement the weighted least squares method and improve the performance of pseudorange based distance detection.

First of all, a satellite is selected from all the candidate satellites as the reference for double difference calculation. As shown in Eq. \ref{eq:noise}, if the variance of $(\varepsilon^0_a - \varepsilon^0_b)$ can be kept small, it can be fairly assumed that the $\mathbf{\epsilon}$'s are uncorrelated and $\mathbf{W}$ can thus be simplified as

\begin{equation}
\mathbf{W} = diag(\frac{1}{(\sigma^1)^2}, \cdots, \frac{1}{(\sigma^n)^2})
\end{equation}

\noindent where $diag(\cdot)$ denotes a diagonal matrix and $\sigma^i$ is the standard deviation of $\epsilon^i$. With independent and zero mean random variables of $\varepsilon^0_a$ and $\varepsilon^0_b$, the only way to keep $(\varepsilon^0_a - \varepsilon^0_b)$ small is to keep the standard deviation of both $\varepsilon^0_a$ and $\varepsilon^0_b$ small. It is thus better to select the satellite with the best CNRs to both of the two receivers as the reference satellite for double difference. To eliminate the impact of noisy measurements from reference satellites, a threshold of $\mathbf{CNR_{ref}}$ is used to set an upper limit to the noise level of the reference satellite. If no candidate satellite reaches the threshold of $\mathbf{CNR_{ref}}$, the whole set of measurements are  dropped and no distance calculation is done for this round. With a selected high-CNR reference satellite, which is indexed as satellite $0$, $\epsilon^i$ can then be approximated as:

\begin{eqnarray}
\epsilon^i \approx (\varepsilon^i_a - \varepsilon^i_b)
\label{eq:noise_2}
\end{eqnarray}

After the successful selection of the reference satellite, the matrix $\mathbf{H}$ and the column vector $\mathbf{D}_{ab}$ are determined. The only unknown left for the left side of Eq. \ref{eq:wls} is  the weight matrix of $\mathbf{W}$. With known CNR $\phi^i_a$ of the received signal from satellite $i$ to GPS receiver $a$, we assume that the variance $(\sigma^i_a)^2$ of $\varepsilon^i_a$ is reversely proportional to the value of $\phi^i_a$ (in dBHz), i.e. $(\sigma^i_a)^2 \propto 1/\phi^i_a$. As $\varepsilon^i_a$ and $\varepsilon^i_b$ are independent from each other, the variance $(\sigma^i)^2$ of each $\epsilon^i$ can thus be derived as:

\begin{eqnarray}
(\sigma^i)^2 &=& (\sigma^i_a)^2 + (\sigma^i_b)^2\\
        &\propto & 1/(\phi^i_a)^2 + 1/(\phi^i_b)^2
\end{eqnarray}

\noindent and the weight matrix $\mathbf{W}$ can then be finalized as

\begin{eqnarray}
\mathbf{W} = diag\left(\frac{(\phi^1_a)^2 \cdot (\phi^1_b)^2}{(\phi^1_a)^2 + (\phi^1_b)^2}, \cdots, \frac{(\phi^n_a)^2 \cdot (\phi^n_b)^2}{(\phi^n_a)^2 + (\phi^n_b)^2}\right)   \nonumber
\end{eqnarray}

Last, in selecting candidate satellites, a threshold $\mathbf{CNR_{min}}$ is used to prevent pseudorange measurements with too much noises from being used in the double difference calculation. Only those satellites with better signal qualities than $\mathbf{CNR_{min}}$ can be selected as candidates for consequent processing.

\subsection{DSRC Assisted Cooperative Distance Detection}

In applying the proposed pseudorange based cooperative distance detection technique to vehicular systems, Dedicated Short Range Communications (DSRC) are needed to exchange raw pseudorange information among neighboring vehicles. The periodical pseudorange measurements can be piggybacked into the DSRC heartbeat messages. As the feasibility of double difference algorithm depends largely on synchronized pseudorange measurements among different GPS receivers, it is recommended that all the GPS receivers conduct their pseudorange measurements with a same fixed time frequency and according to a same fixed time offset. For example, they can conduct their pseudorange measurements at the beginning of every GPS second and output their measurements once every second. As different satellites have different signal propagation time, measurement interpolations may be needed to ensure local synchronization among all the pseudorange measurements of different satellites.

\begin{figure}[htb]
\centering
\includegraphics[width=3.4in, height=0.5in]{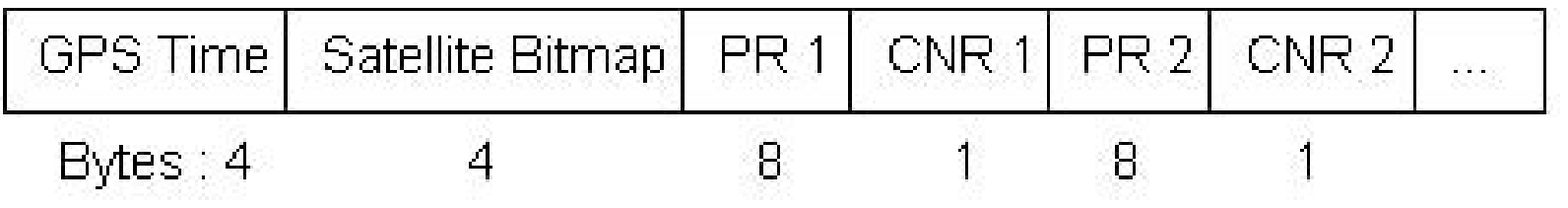}
\caption{Format of the piggyback message}
\label{fig:message}
\end{figure}

Upon finishing each round of pseudorange measurement, the vehicle selects the candidate pseudorange measurements that have better CNRs of the receiving signal than $\mathbf{CNR_{min}}$. Among these candidate pseudoranges, the measurement with the best CNR is picked up. If the number of candidate measurements is less than four or when the best measurement has a CNR less than $\mathbf{CNR_{ref}}$, no information will be piggybacked and broadcasted this round. Otherwise, the GPS time tag and all the candidate pseudorange measurements as well as their CNRs are piggybacked to the end of heartbeat message and broadcasted to all the one-hop neighbors. The format of the attached message is illustrated in Figure \ref{fig:message}.

\noindent The 4-byte Satellite Bitmap field is made up of 32 ON/OFF flags, indicating the presence of the pseudorange measurements of the corresponding satellites in the following message. Each measured pseudorange (PR) is represented by a 8-byte double-precision floating-point data. When there are $n$ candidate satellites, the total length of the attached message is $l = 8+9 \times n$ bytes. With four candidate satellites, which meets the minimum requirement, the attached message is 44 bytes. The same set of information is copied and cached locally together with the output of GPS fix for later processing.

Upon receiving a heartbeat message with piggybacked pseudorange information from a one-hop neighbor, the receiving vehicle checks the GPS time tag  and picks the matching set of measurements from its local cache for processing. Among these two sets of pseudorange measurements, those shared satellites are selected. If the number of shared satellites is less than four, or if there is no satellite satisfies the $\mathbf{CNR_{ref}}$ threshold, the received information is simply dropped without calculating the distance. Otherwise, the satellite with the best CNRs to both of the two receivers is selected as the reference, and the distance is calculated using the proposed algorithm. In building the matrix of $\mathbf{H}$, the cached local GPS fix is used, and the satellite position is calculated using local almanac data.

\section{Experiments and Results} \label{sec:experiment}

To validate the performance of the proposed algorithm, two field experiments are conducted. In these experiments, two GPS receivers with SiRFstar III module are placed on a roof top with distances of three and eight meters apart respectively. The SiRF proprietory binary protocol \cite{Ref:sirf} was used to extract the raw pseudorange measurements, satellite positions, estimated GPS time, as well as the calculated GPS fix. Pairs of independent measurements with the same estimated GPS time are taken out from the two modules for subsequent processing. The CNRs of all pseudorange measurements are also extracted, and only those with CNR greater than  $\mathbf{CNR_{min}}$ are used for distance calculations. The $\mathbf{CNR_{min}}$ threshold is set to 30dBHz, and the $\mathbf{CNR_{ref}}$ threshold is set to 47dBHz.

\begin{figure}[htb]
\centering
\includegraphics[width=3.6in]{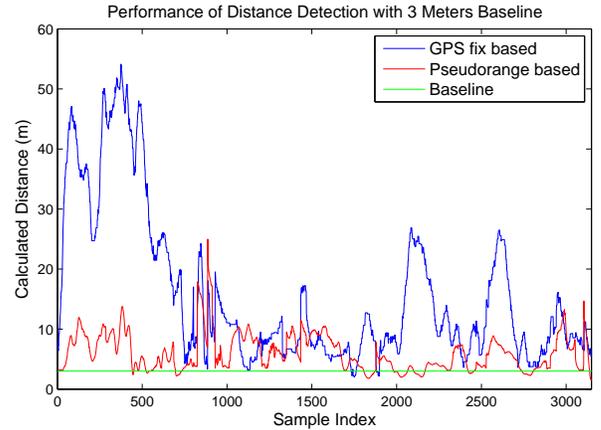}
\caption{Experiment results}
\label{fig:distance_3}
\end{figure}

\begin{figure}[htb]
\centering
\includegraphics[width=3.6in]{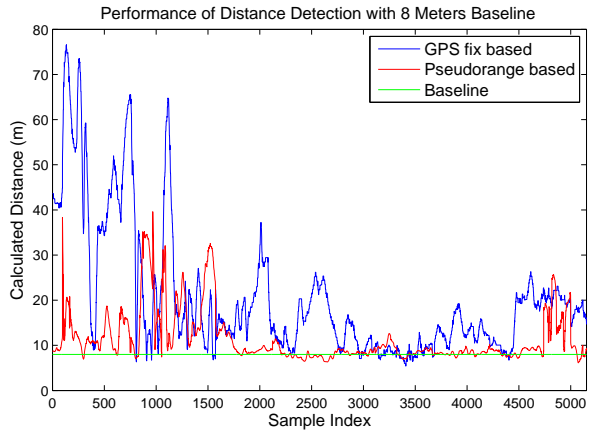}
\caption{Experiment results}
\label{fig:distance_8}
\end{figure}

Figures \ref{fig:distance_3} and \ref{fig:distance_8} show the performance of the weighted least squares pseudorange double difference algorithm in the two experiments. The distance derived by the proposed algorithm is compared with the distance calculated direct from the two GPS fixes generated by the two GPS modules. As demonstrated in the two figures, the weighted least squares double difference method greatly improves the accuracy level of GPS based distance detection. Even when the GPS measurements are severely affected by random noises as shown in the first 1000 samples of the two figures, the proposed algorithm can still provide a decent average noise of less than ten meters, whereas the GPS fix based algorithm goes wrong as far as tens of meters. Although there are still some distance error spikes around twenty meters level, they are unlikely to be directly used for lane level vehicular positioning.  Compared with the distance errors calculated from the GPS fixes, which are around fifty to seventy meters, the proposed solution shows the remarkable performance improvement. Advanced postprocessing algorithm, such as kalman filters, can be used to further improve the performance by filtering out such measurement disturbances.

Table \ref{table_compare} gives an overall performance comparison of different GPS based algorithms in distance detection. There are five algorithms compared, including GPS fix based algorithm, least squares single difference (LS-SD), least squares double difference (LS-DD), weighted least squares single difference (WLS-SD), and weighted least squares double difference (WLS-DD) algorithms. With single difference based algorithms, the difference of clock bias related range bias $(t_a-t_b)$ is set as the fourth dimension of the unknown vector in addition to the three dimensions of the baseline vector $\vec{r}_{ab}$. This unknown vector can be derived in similar ways as with double difference method. It can be solved with either normal least squares based algorithm or weighted least squares based algorithm. For fair comparison, the results of GPS fix based algorithm and single difference based algorithms are averaged over the same set of samples that are eligibly selected by the double difference based algorithms. The average distance error is calculated by:

\begin{eqnarray}
\overline{\Delta d} = \frac{\sum^M_{i=1} | d_i - d_B|}{M}
\end{eqnarray}

\noindent where $M$ is the total number of valid samples, and $d_B$ is the true baseline distance.

As shown in Table \ref{table_compare}, when the GPS fixes are directly used for distance calculation, the average distance errors are more than 12 meters. In contrast, the distance errors are dramatically decreased to less than 7 meters for the other cooperative algorithms where the raw pseudorange measurements are used. In particular, with normal least squares algorithm, the single difference method results in average errors of 5.668 meters and 6.505 meters, which gives around $50\%$ improvements comparing to GPS fix based method. When double difference is used, the average error further decreases by about 1 meter. These improvements are due to the decreased dimension of the unknown variable comparing with the single difference method. When the weighted least squares method is used, which takes into account the signal quality of each pseudorange measurements, the performance is further improved. Even with single difference, the average errors decrease by about 1 meter comparing with using double difference with normal least squares method. When double difference is used together with weighted least squares method, the average errors reach the least values of 3.192 meters and 3.641 meters in the two experiments. The overall performance improvements comparing with GPS fixed based method are more than $70\%$.

As mentioned previously, our proposed algorithm is most
effective when the code acquisition errors are the dominant
errors in the GPS signal. On the other hand, if there is severe
multipath, the performance of our algorithm would be close
to that of the least squares double difference algorithm.

\begin{table}[!t]
\renewcommand{\arraystretch}{1.3}
\caption{Average Distance Errors (m) with Different Algorithms}
\label{table_compare}
\centering
\begin{tabular}{|c||c|c|c|c|c|}
\hline
 & GPS Fix & LS-SD & LS-DD  & WLS-SD & WLS-DD \\
\hline
\hline
3m Baseline & 12.801 & 5.668 & 5.084 & 3.850 & 3.192  \\
\hline
8m Baseline & 12.244 & 6.505 & 5.596 & 3.908 & 3.641 \\
\hline
\end{tabular}
\end{table}

\section{Conclusions and Future Work}
In this paper, we proposed a weighted least squares pseudorange double difference algorithm for cooperative distance measurement in vehicular networks. It makes use of the existing GPS receivers in each vehicle and DSRC-based vehicular networks for accurate distance measurement. With commercially available GPS receivers, our technique can greatly eliminate the impact of those random noise in coarse code acquisition, and reduce the distance estimation error to about 3 meters, which is a great improvement compared with using GPS fixes directly.

The current solution is effective in scenarios where the dominant errors come from the code acquisition. In our future work, the influence from the multipath effect as well as the countermeasures will be further examined. With the improved distance estimation accuracy, we aim to recover the lane level relative relationship among neighboring vehicles by developing advanced cooperative vehicular positioning techniques.


\section*{Acknowledgment}
This research is supported by the Singapore National Research Foundation (NRF) through the Singapore-MIT Alliance for Research and Technology (SMART) Future Urban Mobility (FM) Interdisciplinary Research Group (IRG).



%\bibliographystyle{IEEEtran}

\end{document}